\documentclass[sigchi-a]{acmart}

\def\BibTeX{{\rm B\kern-.05em{\sc i\kern-.025em b}\kern-.08emT\kern-.1667em\lower.7ex\hbox{E}\kern-.125emX}}
    
\copyrightyear{2019}
\acmYear{2019}
\setcopyright{rightsretained}
\acmConference[CHI HCML Perspectives '19]{CHI 2019: Human-Centered Machine Learning Perspectives Workshop}{May 04, 2019}{Glasgow}
\acmDOI{}
\acmISBN{}

\usepackage{pbox}
\usepackage{multirow}
\usepackage{enumitem}
\begin{document}

%
\title{Unfairness towards subjective opinions in Machine Learning}

%
\author{Agathe Balayn}
\email{A.M.A.Balayn@tudelft.nl}
\affiliation{%
  \institution{Delft University of Technology, IBM Center for Advanced Studies}
}

\author{Alessandro Bozzon}
\email{a.bozzon@tudelft.nl}
\affiliation{%
  \institution{TU Delft, Web Information Systems}
  \streetaddress{Van Mourik Broekmanweg 6}
  \city{Delft}
  \country{the Netherlands}}
  
\author{Zolt\'{a}n Szl\'{a}vik}
\email{zoltan.szlavik@nl.ibm.com}
\affiliation{%
  \institution{IBM Center for Advanced Studies}
  \streetaddress{Johan Huizingalaan 765}
  \city{Amsterdam}
  \country{the Netherlands}}

%
\renewcommand{\shortauthors}{Balayn, et al.}

%

%
\begin{abstract}
Despite the high interest for Machine Learning (ML) in academia and industry, 
many issues related to the application of ML to real-life problems are yet to be addressed. 
Here 
we put forward one 
limitation which arises from a lack of adaptation of ML models and datasets to specific applications. 
We formalise a new notion of unfairness as exclusion of opinions. 
We propose ways to quantify
this unfairness,
and aid understanding its causes through visualisation. These insights into the functioning of ML-based systems 
hint at methods to mitigate unfairness. 
\end{abstract}

%

%
\maketitle

\section{Introduction}
\begin{sidebar}
\section{Keywords}
Machine Learning, unfairness, subjectivity, bias, toxicity prediction
\end{sidebar}

Machine Learning (ML) is increasingly employed in real-life applications. Originally used to classify various types of samples based on objective labels, 
it is now also employed for classification tasks of subjective labels (see sidebar)~\cite{o2017weapons,skirpan2017authority}. 
In such tasks, samples can not be associated with clear unique ground truth labels since the property to be assessed (and predicted) is subjective, and might therefore
\begin{sidebar}
\textbf{Examples of objective classification tasks:}
\begin{itemize}
    \item digit recognition from images~\cite{lecun1990handwritten},
    \item human activity recognition from videos~\cite{herath2017going},
    \item spam filtering from text~\cite{almeida2016text}.
\end{itemize}

\textbf{Subjective classification tasks:} 
\begin{itemize}
    \item violence of a video segment~\cite{schedl2015vsd2014},
    \item aesthetic of an image~\cite{bianco2016predicting},
    \item sentiment of a sentence~\cite{tang2015document},
    \item toxicity of a sentence~\cite{wulczyn2017ex}.
\end{itemize}

\textbf{Opinion diversity in subjective tasks:}

\autoref{tab:intro_example} (below) illustrates the diversity of annotators' opinions on toxicity, which would be ignored in a traditional single-label dataset.
61\% of the samples in the dataset bear disagreements for binary annotations, and even more for a 5-point Likert scale.
\end{sidebar}
\begin{margintable}
	\footnotesize
	\centering
	\caption{Example samples and annotations of an ML dataset for the task of predicting sentence toxicity (T: toxic, NT: non-toxic).	}
	\label{tab:intro_example}
	\begin{tabular}{@{}lr@{}}
		\textbf{sample}  & \textbf{annotations}   \\  \midrule[2pt]
		\pbox{0.28\textwidth}{Is there perhaps enough newsworthy information to make an article about the Bundy family as a whole, that the various family members can be redirected to? Or does that violate a guideline I'm not aware of?}
		 & \pbox{0.08\textwidth}{NT(100\%)}
	\\ \midrule[0.5pt]
		\pbox{0.28\textwidth}{What shit u talk to me, communist rat?}  & \pbox{0.08\textwidth}{T(100\%)} 
		\\ \midrule[0.5pt]
		\pbox{0.28\textwidth}{Please relate the ozone hole to increases in cancer, and provide figures. Otherwise, this article will be biased toward the environmentalist anti-CFC point of view instead of being neutral.}
		& \pbox{0.08\textwidth}{T(20\%) NT(80\%)}
		 \\\midrule[0.5pt]
		\pbox{0.28\textwidth}{The article is true, the Israeli policies are killing Arab children.}  & \pbox{0.08\textwidth}{T(50\%) NT(50\%)}
		 \\ \hline 
		 
	\end{tabular}
\end{margintable}
be subject to biases or differences in perception and interpretation 
from each individual assessor of the dataset. 
We argue that this shift of focus has a potential negative impact on the end-users of the applications whose opinions/perspectives might not be reflected by the system when they have not been captured in the data or/and in the model's outputs. Specifically, we claim that the traditional way of conceiving ML-based systems is not adapted to the cases of subjective classifications, and that it leads to unfairness towards certain end-users but also to potential dangers towards society as a whole. Although the impact concerns end-users, we use interchangeably \textit{assessors} and \textit{end-users} since we measure its magnitude based on the assessors' data used as a proxy for the end-users' data that are 
not available in the dataset. The issue concerning differences between the assessors and end-users populations is relevant, but not within the scope of this work. 

ML models usually output a single label per input
representing a single opinion (often the majority one) or the averaged opinion (in case of numerical labels) which might not correspond to any individual. 
Outputting this label (or its distribution) might not be sufficient depending on the 
application. 
First, it conducts to ignoring part of the users' opinions in potentially
different proportions. That makes certain users' experience of the system less valued than for others and can be perceived as unfairness towards these end-users whose opinions might never be showcased. Second, always ignoring certain opinions, mostly from the minority, and accounting for specific opinion trends contribute to the reinforcement of filter bubbles on the Web. This is an emergent danger for societies~\cite{bozdag2015breaking}. 

We name this issue \textit{unfairness as a notion of opinion exclusion} and define it as an \textit{inequality of inclusion of the opinions of the users in the outputs of the ML system}.
We put forward that, by resolving it, it would be possible to minimise potential negative effects coming from the applications of the systems both on a user and society level.
In the following, we propose an initial strategy to evaluate and mitigate this unfairness,
that we expose through the example of an ML model trained to predict whether a sentence is toxic or not.
Yet the work is applicable to any subjective classification task.


\section{Causes of unfairness along the ML pipeline}

\begin{marginfigure}
  \centering
  \includegraphics[width=0.9\linewidth]{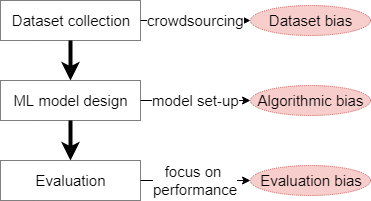}
  \caption{The sources of unfairness in the traditional ML pipeline.}
  \Description{The sources of unfairness in the traditional ML pipeline. We show 3 sources of unfairness in the following order. During the dataset collection part, a dataset bias arises from crowdsourcing. During the design of the ML model, an algorithmic bias arises from the way the model is set-up. During the evaluation part, an evaluation bias arises from the fact that usual evaluation methods only focus on accuracy-related performance.}\label{fig:unfairness_sources}
\end{marginfigure}

Looking deeper into the traditional ML pipeline, we identified three elements which contribute to causing unfairness, that all have to be addressed (\autoref{fig:unfairness_sources}).

\begin{marginfigure}
  \centering
  \includegraphics[width=0.95\linewidth]{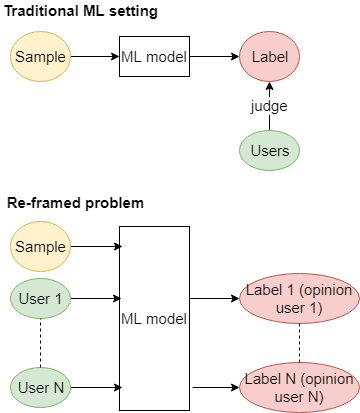}
  \caption{Traditional ML task and re-framed problem. }
  \Description{The traditional and re-framed ML problem. The traditional model takes as input a sample and output a label which is judged correct or not by the users of the model. In the reframed problem, the model takes as input a sample and a user, and output a label which corresponds to the opinion of the specific user on the sample.}
  \label{fig:ML_pipeline}
\end{marginfigure}

\textit{\textbf{Algorithmic bias:}} 
ML research tackling the classification of subjective properties simply considers that the task can be represented using unique binary labels ignoring the subjectivity~\cite{bianco2016predicting,tang2015document} sometimes removing the data with the most disagreement, and aggregating the rest. Soft labels are also employed~\cite{sharmanska2016ambiguity,wulczyn2017ex} to account for the opinion diversity but this does not enable to identify the opinion of each individual.
Only one paper~\cite{binns2017like} has considered the opinions of different categories of population (male and female) but it was shown that there is also disagreement within the categories. 
We suggest the whole problem to be re-framed in order to address unfairness. Here we do not wish to output one label for one sample, but one label for an input being the data sample and the specific user of the application (\autoref{fig:ML_pipeline}). 
\begin{sidebar}
In the re-framed problem, we advocate for outputting one label for each input tuple consisting of a sample and an end-user. During the training phase, this corresponds to tuples of samples and assessors.
\end{sidebar}
In this way, the creators of the application would have access to the opinion of each individual user and they could adapt the decision making process within the system not only to the data sample but to the user as well. For instance in social media, where one would wish to filter out toxic sentences, for each end-user we would not hide information that they, individually, might not see as non-toxic, but they would be protected from what they individually might perceive as toxic (e.g. in cases where a child and an adult do not have the same perception of sentence toxicity).

\textit{\textbf{Dataset bias:}} The way of creating datasets 
tends to bias the data towards certain types of opinions.
Since models are trained on these data, their outputs are automatically unfair. Besides, they are tested on the same type of data and consequently the unfairness might be missed.
Indeed, most ML datasets have their labels collected via crowdsourcing 
(annotators label samples reflecting their opinions on these), which should enable the collection of true opinions. 
However, due to annotation quality issues (mistakes, spamming), researchers and practitioners need to exclude low quality inputs, but current methods can not distinguish these from low popularity opinions, and lead to the selection of only one opinion per sample, leading to exclusion.
Indeed 
to avoid the problem, several annotators label a sample and the annotations are aggregated into a unique, more accurate label (e.g. by majority-voting -MV or by a probabilistic approach~\cite{raykar2010learning,whitehill2009whose}), assuming that the more annotators there are the higher the chance is to get a majority of correct annotations.
This implies that correct labels are labels on which annotators agree, but subjectivity involves disagreement and consequently the assumption does not hold (example \autoref{tab:intro_example}). 
Thus, this process automatically biases datasets towards a unique kind of opinion reflected via the aggregation method selected. 
Few works leverage the disagreement to filter out the wrong annotators and annotations. The CrowdTruth framework enables to compute quality scores to quantify the annotators' quality and the samples' ambiguity based on their disagreement~\cite{aroyo2013crowd}. Collaborative approaches of discussion and argumentation between annotators are also taken to refine the crowdsourcing task and get high-quality labels~\cite{chang2017revolt,drapeau2016microtalk}, but this still implies a unique ground truth. Groups of annotators with similar annotation trends can also be identified by clustering to possibly discover annotators of high-quality or having specific interpretations of the tasks, but this is not yet scalable to a large number of annotators annotating different samples~\cite{kairam2016parting}.

\textit{\textbf{Evaluation bias:}} 
The notion of fairness as exclusion of opinions being new, there is no established way to evaluate it.
Previous research 
is directed at unfairness as a notion of discrimination towards protected categories of population, focusing on models which classify people over certain labels (e.g. whether someone who committed a crime will reoffend)~\cite{gajane2017formalizing}. Recently, research has also studied unfairness in the ML process (mainly the choice of possibly unfair features)~\cite{grgic2018human}. 
Consequently, current systems' evaluation methods are biased towards performance metrics and unfairness as discrimination.
The lack of adapted metric fosters exclusion in current systems since ML practitioners are not aware of the problem, or have no tool to consider it in their systems.
Therefore, we advocate for investigating the evaluation of unfairness and propose initial research ideas in the next sections.

\section{Methods to understand and mitigate unfairness}

\begin{sidebar}
\textbf{Requirements for the unfairness measure:}
\begin{enumerate}
    \item Quantifies unfairness.
    \item Quantifies general model performance (to observe the trade-off between fairness and performance since a model could be totally fair and inaccurate).
    \item Is independent from the evaluation dataset.
    \item Is adaptable to the performance metric(s) important for the application. 
    \item Provides insights into causes of unfairness. 
\end{enumerate}

\textbf{Grouping criteria to understand unfairness.}\\
We assume that the main reason for which unfairness can occur is the difficulty of ML models to predict certain opinions (generally the uncommon opinions). This difficulty might manifest in multiple ways that can be investigated through the following grouping settings.
\begin{itemize}[leftmargin=*]
    \item{Sample-level:} Ambiguous samples might present more disagreement making the opinions on them harder to discover by the model. Grouping samples based on their ambiguity 
    would point out these difficulties.
    \item{Annotation-level:} Popular annotations for a sample are seen more often by the ML model during training, and consequently, should be easier to predict. Grouping annotations based on their popularity (percentage of identical annotations within a sample's annotations) would highlight this.
    \item{User-level:} Apart from the disagreement among users, users of a certain category of population might have similar opinions. Grouping these users would show whether certain categories are discriminated against for the benefit of others. 
\end{itemize}
\end{sidebar}

\subsection{Our proposition to identify and quantify unfairness in ML}

Here we define an \textit{ML model to be unfair when its performance is unequal across its users} (opinions would not be equally accounted for among users). Although this definition suggests easy ways to identify whether a model is unfair, we also need 1) to quantify unfairness in order to minimise it and to compare models on other criteria than traditional performance measures, and 2) to investigate potential causes of unfairness.

To quantify unfairness, a metric should satisfy the requirements listed in the sidebar. We propose the following evaluation method. 
a) Group the users of the model in the dataset based on their disagreement rate with the other users (we compute the average disagreement rate -\texttt{ADR}- per user as the percentage of times a user's annotations are different from the MV because we assume the MV is representative of the outputs of the traditional unfair ML model). This enables to obtain comparable values for different datasets which would be constituted of the same groups' characteristics. 
b) Compute model performance for each user, and the mean of these values within each group. 
c) Compute the standard deviation and mean across the groups' performance to quantify unfairness and general model performance, respectively. The performance metric is chosen by the ML practitioner and several metrics can be combined by averaging the results. 

To investigate potential causes of unfairness, we visualise the performance of each individual group. We hypothesize it enables to identify where inequalities come from and to point out the types of users for which the predictions should be improved.
Additionally, we suggest other criteria (see sidebar) to group the dataset in order to better understand how the ML model behaves and which are the inaccuracies that make the model unfair.

\subsection{Considerations to decrease unfairness}
To mitigate dataset bias and maintain a diversity of opinions, we must not aggregate the annotations. 
We propose first to use a quality control mechanism (e.g. CrowdTruth framework) to filter out the lowest quality annotators giving wrong annotations. Second, we can employ disagreement as a signal to identify valid but unpopular opinions and differentiate them from occasional annotators' mistakes. 

Regarding algorithmic bias, we envision two main lines of work which could eventually be combined. Usually, few data points per user are known at training time, thus it is neither possible nor scalable to train one accurate model per user. A solution is to train models with certain parameters conditioned on the users (e.g.~\cite{tang2015user} for recommender systems), where preferences of new users should be learned at run time. Another possibility is to leverage the knowledge of other fields such as Psychology to find the internal characteristics of a user, which influence their perception of a label (or deduce these variables with additional ML models) and input these as features encoding each user.

\subsection{Examples for sentence toxicity prediction}

\begin{sidebar}
\textbf{The three ML models:}
\end{sidebar}

\begin{margintable}
	\footnotesize
	\centering
	\caption{First two ML models used to analyse the unfairness evaluation method.
	}\label{tab:unfairness_models}
	\begin{tabular}{lcc}
		 & \textbf{\textit{Model 1}} & \textbf{\textit{Model 2}} 
		 \\  \midrule[2pt]
		\pbox{0.06\textwidth}{Inputs}
		 & samples & \pbox{0.15\textwidth}{samples + demographics} 
	\\ \hline
		\pbox{0.15\textwidth}{Ground truth}  & MV & annotations 
		\\ \hline 
		\multicolumn{3}{l}{\textbf{\textit{Unfairness value}}}   \\
		 & {0.07} & 0.04 
		 \\ 
		\multicolumn{3}{l}{\textbf{\textit{General performance}}}  \\
		 & {0.68} & 0.68 
		 \\ \hline \\
		 
	\end{tabular}
\end{margintable}

\begin{sidebar}
We add the demographic information to \textit{Model 2}'s inputs because Psychology literature highlights these variables as the most influencing variables for perception of sentence toxicity~\cite{cowan1996judgments}. 
\textit{Model 3} is a hypothetical, perfectly accurate model which returns the exact annotations for each annotator, and consequently, is fair.

We instantiate the models with Logistic Regression classifiers 
trained and tuned using 5-fold cross validation, and compute the accuracy performance on a balanced evaluation dataset.
\end{sidebar}

\begin{marginfigure}
\centering
  \includegraphics[width=0.8\linewidth]{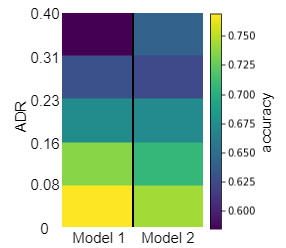}
  \caption{Visualisation of the unfairness based on groups of annotators' average disagreement rate with the MV (ADR).}
  \Description{Visualisation of the unfairness based on groups of annotators' average disagreement rate with the MV (ADR). In each row we see the performance accuracy of the two models on the different group of annotators. Model 2's performance are more equal across rows than model 1's performance.}
  \label{fig:unfairness_visualization}
\end{marginfigure}
We illustrate our method for a better understanding of unfairness using the dataset of~\cite{wulczyn2017ex} for sentence toxicity prediction, which includes opinions of ten annotators per sample with corresponding demographic information (age, gender, education). We build three ML models with three different unfairness-related behaviors (see sidebar). \textit{Models 1 to 3} are expected to be less to most fair due to their training process. 
\autoref{fig:unfairness_visualization} shows an example visualisation of the groups' performance for annotators grouped on ADR scores. As expected, \textit{Model 1} is more unfair than \textit{Model 2} since its performance across groups is more disparate than for \textit{Model 2}. The perfect model would present equal accuracy of 1 for each group and consequently an unfairness of 0 and performance of 1. The unfairness score (\autoref{tab:unfairness_models}) reflects this trend with a higher value for \textit{Model 1}, while the general performance is equal for the two models.
The visualisation furthermore confirms the sources of unfairness in the models. We observe that \textit{Model 1} is inaccurate at predicting the opinions of users who tend to disagree with the MV the most; this is better in \textit{Model 2}, but not fully solved. ML practitioners could use these observations to further improve their model, for example, by collecting more data from specific annotators.  

\subsection{Discussion}
The evaluation method meets the requirements to better understand unfairness.
Although applied to a specific case focusing on accuracy with Logistic Regression, it is applicable to any type of subjective classification task, any performance metric(s) as well as ML model. To thoroughly evaluate the effectiveness of the method in the future, we recommend first to perform user studies to understand whether the visualizations are both clear and informative to the users, and second to investigate whether integrating the visualizations within a human-in-the-loop unfairness mitigation methodology enables to effectively remove the biases.
Our proposition accounts for group fairness. However, there might exist unfairness inside the groups themselves, and thus, metrics to measure individual fairness would also give different insights into the problem.
Unsupervised clustering of sentences with similar content, or of annotators with similar annotation trends could enable the discovery of other reasons for unfairness. 
E.g. it could identify annotators with a certain opinion trend, which might always be served badly by a model compared to annotators of another opinion line.
Besides, we employed users' demographics to improve classification
, however, this might be a privacy sensitive point for certain applications. Here appears a trade-off not only between fairness and accuracy, but also privacy, that presages more complexity in the ways to mitigate the issue.

\section{Conclusion}

In this paper, we discussed the issue of unfairness in the context of ML-based systems, and we argued for a reconsideration of the problem at hand in other terms with more attention towards the human user. We proposed ways to quantify unfairness, to explore its causes, and to mitigate them.
The proposed approach 
is currently being implemented within the IBM AIF360 toolkit~\cite{aif360-oct-2018}, in order to make it easily available for ML practitioners who would implement systems with potential ethical issues.
This work falls within the broader research direction of ethics, ML and explainability.
Methods to make ML more fair might trigger additional ethical issues that remain to be investigated.

\bibliographystyle{ACM-Reference-Format}
\bibliography{sample-sigchi-a}
\end{document}